\documentclass[conference]{IEEEtran}
\IEEEoverridecommandlockouts
\usepackage{cite}
\usepackage{amsmath,amssymb,amsfonts}
\usepackage{algorithmic}
\usepackage{graphicx}
\usepackage{textcomp}
\usepackage{xcolor}
\graphicspath{ {images/} }
\usepackage{bm}
\usepackage{multirow}
\usepackage{comment}
\usepackage{url}

\newtheorem{lemma}{Lemma}
\newtheorem{theorem}{Theorem}

\def\BibTeX{{\rm B\kern-.05em{\sc i\kern-.025em b}\kern-.08em
    T\kern-.1667em\lower.7ex\hbox{E}\kern-.125emX}}
\begin{document}

\title{Boilerplate Detection via Semantic Classification of TextBlocks
}

\author{\IEEEauthorblockN{Hao Zhang}
\IEEEauthorblockA{\textit{Department of Computer Science} \\
\textit{University of Massachusetts, Lowell, USA}\\
Hao\_Zhang@student.uml.edu}
\and
\IEEEauthorblockN{Jie Wang}
\IEEEauthorblockA{\textit{Department of Computer Science} \\
\textit{University of Massachusetts, Lowell, USA}\\
Jie\_Wang@uml.edu}
}

\maketitle

\begin{abstract}
We present a hierarchical neural network model called SemText to detect HTML boilerplate based on a novel semantic representation of 
text blocks.
We train SemText on three published datasets of news webpages and fine-tune it using a small number of development data in CleanEval and GoogleTrends-2017. We show that SemText achieves the state-of-the-art accuracy on these datasets. We then demonstrate the robustness of SemText by showing that it also detects boilerplate effectively on out-of-domain community-based Q\&A webpages.

\end{abstract}

\begin{IEEEkeywords}
content extraction, sequence labeling, boilerplate detection, word embedding, neural networks
\end{IEEEkeywords}

\section{Introduction}
How to detect HTML boilerplate accurately and efficiently is a continuing quest for search engines and other applications. Methods that worked well on earlier webpages may not work well on contemporary webpages due to new structure and presentation style.

\begin{figure}[h]
\centering
\includegraphics[width=0.50\textwidth, height=3in]
{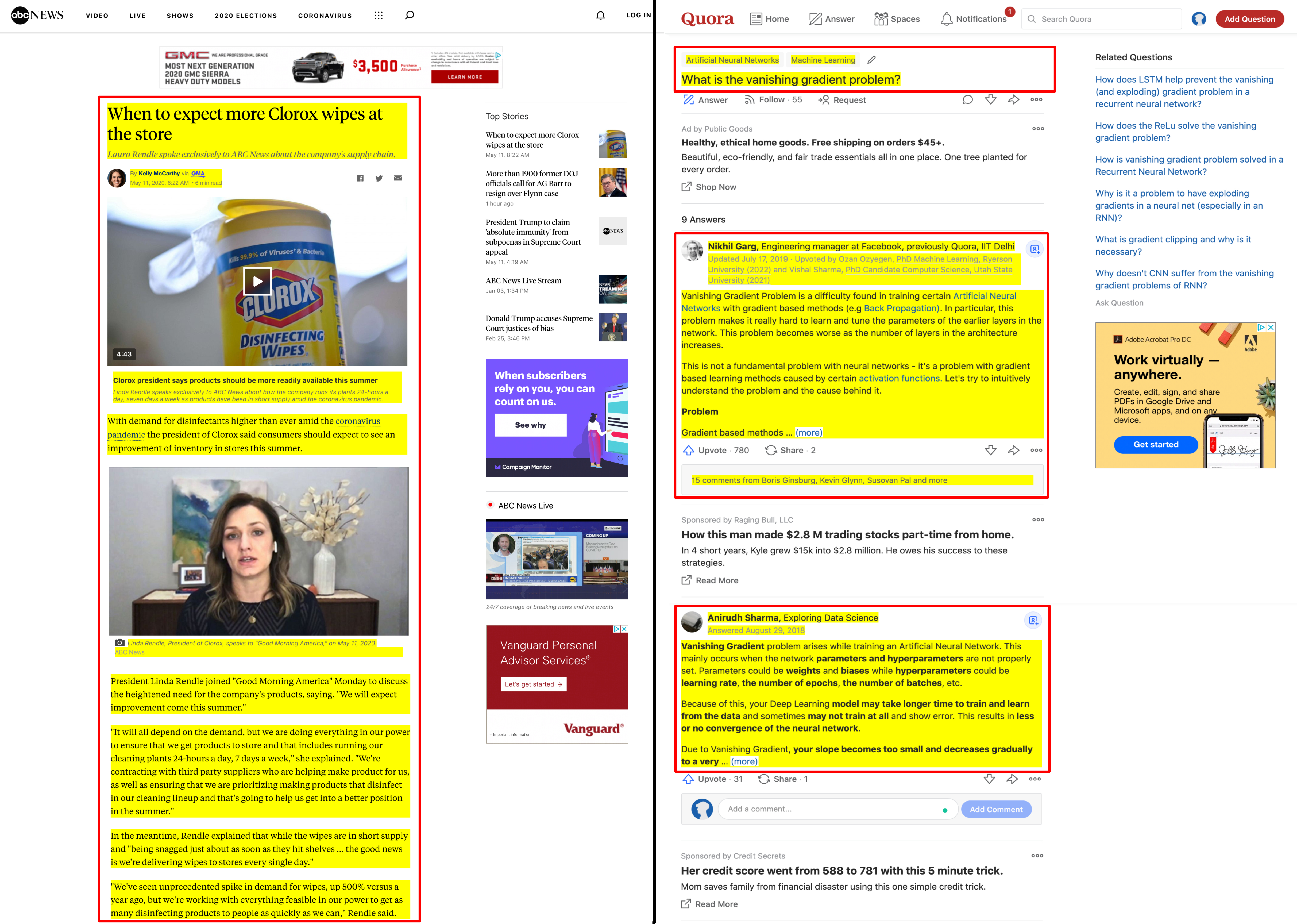} 
\begin{tabular}{cc}
\end{tabular}
\caption{Sample type-1 (left) and type-2 (right) structures, where highlighted texts enclosed in the added red boxes are the corresponding main content}
\label{fig:1}
\end{figure}

Structures of contemporary webpages have two types: The type-1 structure keeps the main content in one place, surrounded by boilerplate contents, while the type-2 structures scatters the main text across the entire webpage. 
News articles and blogs are type-1, and community-based Q\&A webpages are type-2, which display lists of questions followed by one or more answers to each question (see Fig. \ref{fig:1}).

Constructing a spiderbot (a.k.a. spider or crawler) for a given webpage is the most
accurate method to extract the main text. However, it involves tedious manual inspection of the layout of the page, and its accuracy
is sensitive to even the slightest change of the underlying structure of the webpage. To avoid laborious construction and maintenance of spiders, researchers have pursued a different approach known as boilerplate detection 
based on common features, which are less sensitive to structural change of webpages. 

Boilerplate-detection methods have mainly targeted at type-1 pages;
extracting main contents from type-2 webpages has not attracted much attention.
However, there is a need to extract question-answer pairs from community Q\&A websites for
constructing a domain-specific chatbot (e.g., \cite{Yang-Wang2017}). Recent success of neural-network research has made it possible to construct a boilerplate detection model to handle webpages of both types. We present such a model called SemText that requires no handcrafted features, and demonstrate that it is a promising approach.

SemText is a hierarchical neural-network model based on a semantic representation of text. To obtain this representation we partition the text contained in an HTML file into a sequence of text blocks, and represent each text block uniquely with three sequences: (1) a sequence of HTML tags on the path from the root of the DOM-tree of the page to the text block, (2) a sequence of class names on the path from the root to the text block, and (3) a text string (content) enclosed in the text block. HTML tags have semantic meanings, so do class names if web developers follow the conventional naming rules. We take advantage of these meanings by replacing each HTML tag with the underlying word and each class name with an appropriate sequence of words. We view a word string as a sentence with certain meanings, and so semantic representations of words such as word-embedding vectors may be used.
Next, we feed the resulting strings to a depthwise CNN model to produce a feature map of a text block. 
We then apply a Bi-LSTM-CRF sequence-labeling model to classify feature maps using the semantics registered inside and between them.

We train SemText in an end-to-end fashion on a dataset that combines three published type-1 webpages collected from 2008 to 2020, then fine-tune it using a small number of cross-domain webpages for development provided by CleanEval \cite{baroni2008cleaneval} and GoogleTrends-2017 \cite{leonhardt2020boilerplate}. We show that 
SemText achieves the highest F1 score over previous methods on these datasets. We then evaluate SemText on a collection of type-2 webpages and show that SemText also works well. 

\section{Related Work}\label{sec:2}
Previous boilerplate-detection methods can be divided into rule-based, website-based, and machine learning algorithms.

\subsubsection{Rule-based and heuristic methods} They
are based on shallow features such as text length or text-over-tag density. 
For example, the main content could be located at a ``plateau" area where 
the amount of text increases dramatically while 
the number of tags only changes a little \cite{finn2001fact}. The main content could also be in an area with the highest text density defined by the number of tokens over the number of tags or lines \cite{kohlschutter2008densitometric,weninger2010cetr,sun2011dom,wang2015qread}. These methods, while achieving high accuracy on certain webpages such as news articles, do not work well on contemporary webpages with running ads separating the main text into multiple segments. Contemporary webpages would present multiple ``plateaus" or multiple clusters of the same highest density. 

\subsubsection{Website-based methods} They are based on the observation that webpages on the same website would follow the same style. For example, merging the DOM-trees of single webpages on the same website in a top-down fashion, the main content could be a sequence of nodes with more presentation styles and more diverse content types \cite{yi2003eliminating}. A boilerplate may also be viewed as a common subtree of DOM-trees of the webpages at the same website, and so boilerplate detection is reduced to finding a common subtree from a set of given trees \cite{vieira2006fast}. 
These methods can detect website templates with high accuracy, applicable to both type-1 and type-2 webpages. In reality, however, most applications only have a single webpage as a data source without 
the needed website information. 

\subsubsection{Machine-learning methods} They can be categorized 
into regression, classification, and sequence labeling.

Regression methods rank or score text blocks. For example, using a vision-based page segmentation method \cite{cai2003vips} to visually separate a webpage into several blocks by the center pixel coordinate, width, height, and other visual information, a radial-basis-function network can perform regression to score text blocks \cite{song2004learning}. However, most published datasets do not provide ranking or scoring of extractions to support regression, which hinders this approach.

Classification methods represent and classify each text block independently
based on handcrafted features. For example, a total of 67 handcrafted textual features are used 
for block classification \cite{kohlschutter2010boilerplate}. Similarily, FIASCO\cite{bauer2007fiasco} represents a text block using linguistic, structural, and visual features, and uses SVM for classification. However, classifying text blocks without considering context or the interrelations between them is error-prone, especially those with limited content. For example, to classify a text block that says ``limited time 20 percent off" without context is difficult, for it could be a boilerplate or the main text depending on the context.

Sequence-labeling methods, such as Conditional Random Field (CRF) \cite{spousta2008victor,neunerdt2015enhanced}, use relations between neighboring text blocks to jointly decode the best chain of labels for text blocks. 
To enrich conventional features, a CNN model may be added to sequence labeling. For example, Web2text \cite{vogels2018web2text} applies CNN to learn unary and pairwise classification potentials for the sequence of text blocks, maximizing the joint probability using the Viterbi algorithm. But the CNN model is used to refine 128 hand-crafted features instead of learning feature representations. 

Machine-learning methods rely on handcrafted features, ranging from textual to structural and from linguistic to visual; handcrafted features are sensitive to structure change.
Moreover, most handcrafted features were designed specifically for type-1 webpages with simpler layouts. 
To avoid handcrafted features, deep learning models have recently been used to detect boilerplate. BoilerNet \cite{leonhardt2020boilerplate}, for example, is a neural sequence-labeling model that represents each text block as a vector, encoding both HTML tags and words in the text block. Each index in the vector indicates the token count for a specific tag or word. A bi-directional LSTM model is adopted for sequence labeling. BoilerNet outperforms or matches previous models. 

How to represent a text block is critical. Counting tokens may work for HTML tags, for the number of tags is limited. But it may not work well for words. In particular, if a token in the testing webpages has never been seen in the training set, the vector fails to represent the text block properly. 
Although it is possible to represent text with limited tokens by carefully selecting them, it is akin to using handcrafted features. Furthermore, a counting-based representation discards the most important semantic information registered inside the text. Such information is often used by humans when distinguishing the main content from boilerplate.

\section{SemText}\label{sec:4}\label{sec:4.1}

We model boilerplate detection as a text-sequence labeling problem, where a webpage $X$ is divided into a sequence of text blocks $X =[x_1, x_2,\ldots, x_m]$ and each text block is encoded using semantic representations of three word strings: (1) a sequence of 
HTML tags, (2) a sequence of class names, and (3) a sequence of text in the text block. We seek to produce a globally-optimal label sequence $Y = [y_1, y_2, \ldots, y_m]$ for $X$, where $y_t$ denotes the classification label (``main" or ``boilerplate") assigned to block $x_t$ at time $t$ on input $x_t$.

\subsection{Generation of text block sequence}

We present an algorithm called Search-and-Combine Segmentation (SCS) to generate a sequence of text blocks 
 in two phases: the search phase
and the combine phase. In the search phase,
SCS
traverses the DOM-tree of a given HTML file using depth-first search (DFS)
and identifies 
leaf nodes that contain text. 
On each path during the search phase, SCS removes text-formatting tags and the corresponding closing tags, but not the enclosed text. In the combine phase, starting from the first text block, SCS recursively compares the current text block with the next text block and merge them into a new block if they are siblings and with the same presentation style. 
Merging text blocks provides needed
semantic information for more accurate classification and helps to prevent vanishing gradients during training.

\subsubsection*{The search phase} HTML tags are categorized into three groups. 
Tags in group-1 and the enclosed text (if any) do not contribute to the main text. These tags include all document-metadata tags (e.g., $\langle$head$\rangle$, $\langle$meta$\rangle$), all scripting tags, almost all content-embedding tags (e.g., $\langle$img$\rangle$, $\langle$audio$\rangle$), all form tags, among a few others. Tags in group-2 contain text that may or may not contribute to the main content, but the tags should be excluded from a text block. 
These include most of the text-formatting tags (e.g., $\langle$em$\rangle$, $\langle$strong$\rangle$), and some tabular-tags ($\langle$tbody$\rangle$, $\langle$thead$\rangle$). Tags in group-3 and the enclosed text should all be included in a text block. These include all section tags, some grouping tags (e.g., $\langle$div$\rangle$, $\langle$p$\rangle$), all list tags, some tabular tags (e.g., $\langle$table$\rangle$, $\langle$caption$\rangle$), and a few others.

Let $f$ be an HTML file. SCS($f$) performs a depth-first search (DFS) on $f$'s DOM-tree, finding a path from the root node to each leaf node. For each path during search, remove every tag in group-1, together with its closing tag (if any) and the content enclosed. Also removed is every tag in group-2 and its closing tag (if any), but not the enclosed text.

SCS($f$) creates a text block when it encounters a
group-3 tag, opening or closing, 
by extracting the text between this tag and the next group-3 tag, opening or closing; and stores it with the sequence of opening tags and the sequence of class names on the path from the root to this block registered by the DFS search. 
Remove closing tags. 
If the text extracted is empty or invalid,  do not create a text block. Thus, the sequence of text blocks produced by SCS$(f)$ in the order of DFS-traversal is a partition of the valid text portion of $f$ in its original order, where
each text block $x$ consists of three strings of text: $\mbox{tag}(x)$, $\mbox{class}(x)$, and $\mbox{text}(x)]$, with tag($x$) being the sequence of the group-3 tags on the path from the root to the block, class$(x)$ the sequence of class names on the same path, and text$(x)$ the text enclosed in $x$. 

Given a character string $s$, denote by $|s|$ the number of characters contained in $s$. Let $q$ be the height of the DOM-tree for $f$. For a  well-balanced DOM-tree, we have $q = O(\log |f|)$. Without loss of generality, assume that $q \leq O(\sqrt{|f|})$ for any $f$. All webpages we have encountered satisfy this property. Note that the length of each tag and each class name is bounded above by a constant. The following lemma is straightforward.

\begin{lemma} \label{lemma:1}
Let $x_1, x_2, \ldots, x_r$ be the sequence of text blocks generated by
SCD($f$).  We have $\sum_{i=1}^r |\mbox{tag}(x_i)|+|\mbox{class}(x_i)| \leq O(|f|)$
and $\sum_{i=1}^r |\mbox{text}(x_i)| < |f|$.
\end{lemma}

\subsubsection*{The combine phase} 
Let $L$ be an array of text blocks created in the search phase in the order of DFS-traversal, and $L.i$ the $i$-th text block in $L$. Then SCS$(f) =c(L.1,L.2)$, where $c(x,y)$ is defined recursively as follows, and $x$ is stored right before $y$:

\subsubsection*{Case 1}  $x$ and $y$ are leaf siblings with the same tag sequence and the same class-name sequence. Then $x$ and $y$ belong together.
That is, $c(x,y)=z$ is a new block  
with $\text{tag}(z) = \mbox{tag}(x)$, $\text{class}(z) = \mbox{class}(x)$, 
and $\text{text}(z) = \mbox{text}(x) \mbox{text}(y)]$. 

\subsubsection*{Case 2}
$y$ is a leaf and a single child of $x$, class($x$) = class($y$), tag($x$) is a prefix of tag($y$), and $|\mbox{tag}(x)| = |\mbox{tag}(y)|-1$. They $x$ and $y$ belong together.
That is ,combine them into a new text block $c(x,y)$ as in Case 1.

\subsubsection*{Case 3}
$y$ is a leaf, $x$ is either a sibling of $y$ or a parent of a single child $y$, then don't combine. That is, $c(x,y) = \langle x, y\rangle.$ This ``don't-combine'' rule can be justified as follows: If $x$ and $y$ are leaf siblings with $\mbox{tag}(x) \not= \mbox{tag}(y)$, then the two blocks may have different structures. 
If
$x$ and $y$ are leaf siblings with $\mbox{class}(x) \not= \mbox{class}(y)$, then the two blocks may have different presentation styles. If $y$ is a leaf single child of $x$, but $\mbox{class}(x) \not= \mbox{class}(y)$, then the child would have a different intention with the parent. 
\subsubsection*{Case 4} None of the above and $y$ is not the last stored. Let $z$ be its next text block in the order of DFS-traversal. Then let $c(x,y) = c(x, c(y, z)).$

\begin{theorem} \label{thm:1}
SCS$(f)$ runs in $O(|f|)$ time.
\end{theorem}

\textit{Proof sketch.} In the search phase, DFS runs in linear time in terms of the number of nodes of the DOM-tree for $f$, which is determined by the number of opening tags contained in it.Let $x_1, x_2, \ldots, x_r$ denote the number of the temporary text blocks produced by the search phase. While $r$ is much smaller than $|f|$, the search phase may still need to read through the entire file. Thus, it follows from Lemma \ref{lemma:1} that the search phase incurs a running time $O(|f|)$. The combine phase is a linear recursion on non-overlapping text blocks,and so its running time is $O(\sum_{i=1}^r |x_i|) \leq O(|f|)$.
\hfill$\square$

\subsubsection*{Remark} It is reasonable to assume that 
a text block generated during the search phase will not contain both main and boilerplate contents; otherwise, even humans cannot distinguish the main text from boilerplate without inspecting the semantics of the corresponding text, which defies the purpose of the webpage. Likewise, the combine phase will in general not combine a text block for boilerplate and a text block for the main content, because to combine them, they would be siblings with the same tags and class names, which violates the general practice that the main text and boilerplate are distinguishable by humans based on their looks. Our experimental results affirm this observation.

\subsection{Word replacement for tags and class names} \label{sec:3.2}
 
\begin{figure}[h]
\centering
\includegraphics[width=0.5\textwidth]{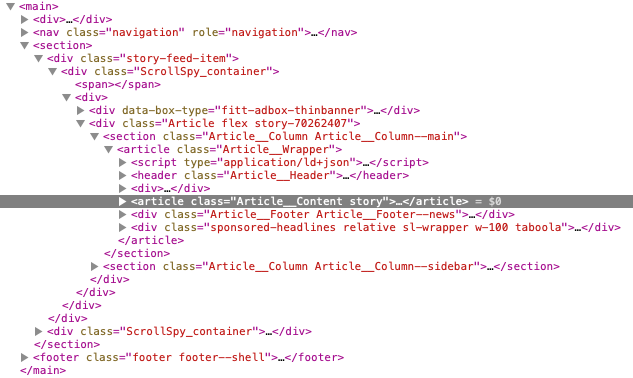}
\caption{The DOM-tree of an ABC news webpage. The darken part represents 
text blocks for the main content 
}
\label{fig:2}
\end{figure}

We observe that using semantics contained in the text enclosed in a text block can help identify boilerplate. For the text enclosed inside a text block, if ``The US president ..." is in one block and ``limited time 20\% off" in another, then it is easier to tell that the former text block is
part of the main content and the latter a boilerplate 
according to their meanings.
To bring in semantics to a text block, we replace each tag with the underlying word or phrase to produce a ``sentence" of words. For example, $\langle$p$\rangle$ is replaced by ``paragraph", $\langle$div$\rangle$ by ``division", and $\langle$h1$\rangle$ by ``primary headline".
Similarly, class names, especially in HTML files that follow conventional naming rules for understanding and reusing codes are often formed by words or meaningful word abbreviations. Fig. \ref{fig:2} is an example of the DOM-tree for an ABC news webpage, where the main text---the darkened part with the news text omitted---has a sequence of class names like, after text cleansing, ``story feed item scrollspy container article flex story article content story". For the blocks around the main content, we can see class names with words of ``header", ``footer", and ``sponsored headline". 

We remove hyphens, dashes, and other delimiters from class names, separate camel-case words, discard numbers, and expand abbreviations to their original words. A class name that cannot be converted to a word string (note that this is rare) is left as is.
We recommend keeping these class names to prevent information loss. 
Another thing worth mentioning is that the ID attributes may be added to the class-name sequence to form an ID-class-name sequence, as used by CSS for manipulating the element with a specific ID. ID-class-name sequences are treated in the same way as class-name sequences.
\begin{figure}[h]
\centering
\includegraphics[width=0.45\textwidth]{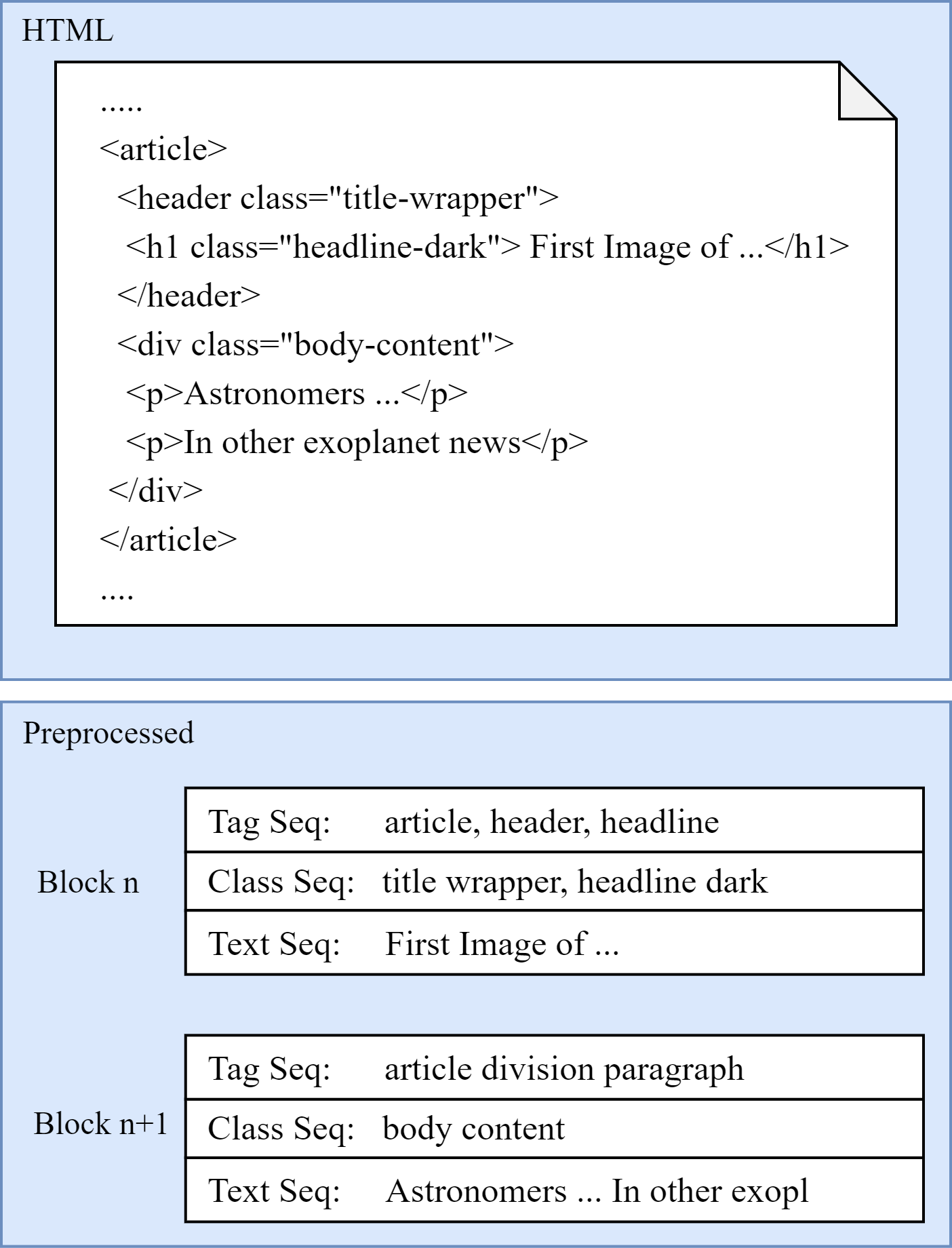}
\caption{An example of representations of word strings for two neighboring text blocks}
\label{fig:3}
\end{figure}
Fig. \ref{fig:3} depicts an example of converting tag sequences and class-name sequences into string of words for two neighboring text blocks.

\subsection{Sequence labeling}
\begin{figure*}[t]
\centering
\includegraphics[width=\textwidth, height=2in]{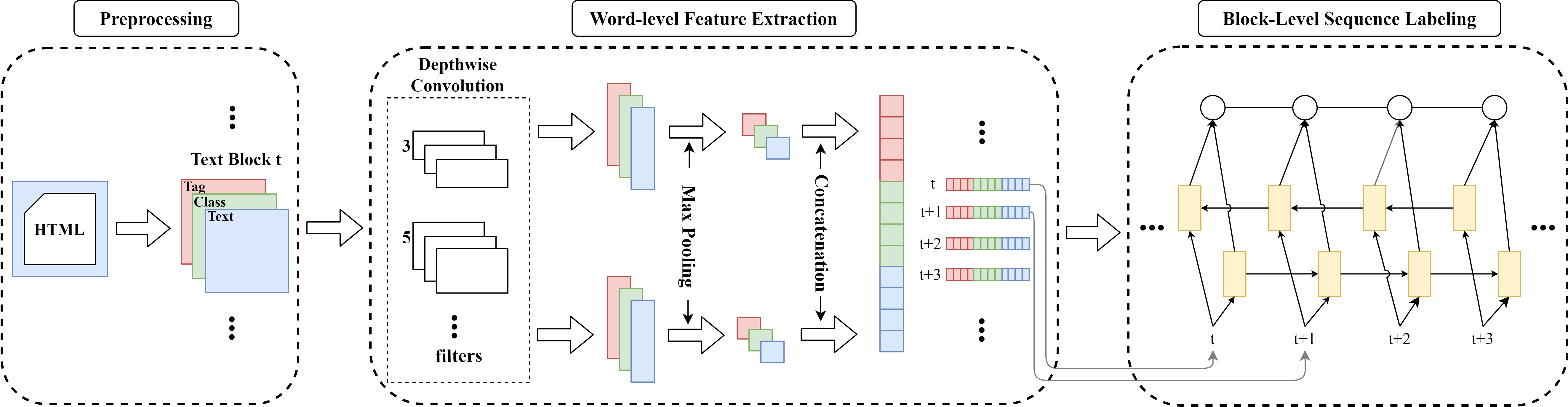}
\caption{SemText architecture}
\label{fig:4}
\end{figure*}

Figure \ref{fig:4} depicts the architecture of SemText. The preprocessing component is responsible for generating text blocks 
from a given HTML file. The word-level-feature-extraction component is a depth-wise CNN model that takes text blocks  
as input and generates a feature map for each text block. The block-level-sequence-labeling component is a Bi-LSTM-CRF model that labels each feature map as ``main" or ``boilerplate".

\subsection{Preprocessing}
\subsubsection*{Word embedding} \label{sec:4.1}
%
SemText removes stop words from each word string obtained from Section B above, and  replaces each remaining word with its word-embedding representation to bring in useful semantic and syntactic information. If a word does not appear in the pre-trained dataset, such as the unidentifiable class names aforementioned, SemText obtains an embedding representaton for it using Subword \cite{bojanowski2017enriching}, a method that overcomes this problem.

\subsubsection*{Text-block truncation}
Long text blocks are undesirable because of intense computation needed to process them. On the other hand, it is not necessary to have long text blocks for the purpose of classifying them. What is needed is a reasonable amount of text in each word string. Thus, for each text block, SemText keeps the first $n$ (e.g., $n = 50$) words (after removing stop words) in each word string and removes the rest.

\subsubsection*{Preprocessing time complexity}
Let $f$ be an HTML file for SemText.   
Generating a sequence of text blocks from $f$ using SCS takes $O(|f|)$ time (Theorem \ref{thm:1}). The time for carrying out word replace and replacing a word with its word-embedding vector is constant. Truncating a text block $x$ is $O(|x|)$ time. Thus, SemText runs in $O(|f|)$ time for preprocessing. 

\subsection{Word-level depth-wise CNN encoding}
We construct a CNN encoder to extract word-level features, motivated by Kim's one-layer CNN model with 1-dimensional convolution 
\cite{kim2014convolutional}. An input $x_t$ has three word strings with at most $n$ words in each string, denoted by $x_t = [x_{t1}, x_{t2}, x_{t3}]$, where each word is a 
$k$-d embedding vector. 
If $x_{ti}$ ($i=1,2,3$) has less than $n$ words, then fill in the rest of the word slots by 0-vectors. Thus, each $x_{t,i}$ becomes an $n \times k$ matrix $\bm{M}_i = [w_{i1}, w_{i2}, \ldots, w_{in}]^T \in \mathbb{R}^{n \times k}$, where $i$  is the feature map number and $w_{ij}$ a $k$-d vertical vector.

Unlike regular CNNs 
that apply convolution filters to all feature maps 
to form a single result, our CNN model uses different feature maps to represent different aspects of a text block not necessarily having any cross-feature-map relation with others, and uses different convolution filters on different feature maps. Thus, we perform depthwise 1-d convolution instead of combining them. An added benefit of doings so is that it significantly reduces the number of parameters. In particular, our depthwise 1-d convolution first splits the input to form different feature maps, then carries out convolution operations on each feature map 
$\bm{M_i}$ with a filter $\bm{W}_{i} \in \mathbb{R}^{v \times k}$, where $v$ is the filter width. Let $\bm{M}_{i(j:j+v-1)}$ denote a filter-size block that contains $m$ word-embedding vectors, then the corresponding feature map for the current window is $c_{ij} = \bm{W}_i^T \bm{M}_{j:j+v-1} + b_{ij},$  where $b_{ij}$ denotes a bias factor. Sliding the filter in the vertical direction, we obtain a feature map as follows: $c_i =[c_{i1}, c_{i2}, c_{i3}, \ldots, c_{i(n-v+1)}].$ Convoluting each feature map with the corresponding filter, and then perform max-pooling over each feature map that extracts the largest number in the map, resulting in a 3-tuple of numerical values for $x_t$. ;

Repeat the above procedure for $\ell$ different filters with different kernel sizes to produce $\ell$ 3-tuples of numerical values. Group them according to feature maps 
and flatten the results into a 1-d output, denoted by $\hat{x}_t$, to produce the final feature map of $x_t$. Figure \ref{fig:4} illustrates the data flow of this process in the word-level feature extraction module.


\subsection{Block-level Bi-LSTM-CRF sequence labeling}
At the block-level, SemText uses the Bi-LSTM-CRF \cite{huang2015bidirectional} architecture to capture context for the block sequence. To learn a contextual representation $\hat{h}_t$ and a context dependency $\hat{c}_t$, the feature map $\hat{x}_t$ generated by the CNN model at time $t$, as well as the previous block information $\hat{h}_{t-1}$ and $\hat{c}_{t-1}$, are fed to a long-short-term memory (LSTM) unit \cite{hochreiter1997long}. LSTM units are connected forwardly, thus previous block information can be captured, which may impact the representation and output for the current input block. LSTM is designed to resolve the gradient vanishing problem that can be encountered when training a traditional recurrent-neural-network model, allowing long-term dependencies being captured. It uses a set of gates to decide when a certain dependence should be remembered, forgotten, or outputted. 

LSTM, however, has a forward-only direction capable of capturing the past context only. To make use of forward and backward information, bidirectional LSTM (Bi-LSTM) \cite{graves2013speech} puts two LSTM together, one in the forward direction and one in the backward direction. Thus, the contextual representation of the $t$-th text block $x_t$  can be rewritten as $\hat{h}_t = [\overrightarrow{h_t}; \overleftarrow{h_t}]$. To make effective use of past and future prediction results, a CRF layer is added to the output of Bi-LSTM for jointly decoding the best chain of labels. Let $H$ denote the contextual representation generated by Bi-LSTM through time, that is, $H = [\hat{h}_1, \hat{h}_2, \ldots, \hat{h}_m].$ The CRF layer maximizes conditional probability $p(Y|H) = \exp({\mbox{score}(H, Y)})/\sum_{Y'} \exp({\mbox{score}(H, Y')})$ to find the highest-scored result over all possible label sequences, where $Y'$ represent a possible sequence and $\mbox{score}(H, Y) = \sum_{i=1}^m \log s(y_{i}, H) + \log t(y_{i-1}, y_{i}, H)$ is a log potential function with $s$ being an emission potential and $t$ a transition potential. The emission potential is defined by a fully-connected layer, transferring the output of Bi-RNN at timestamp $t$ to a score. The transition potential denote a transition probability from the previous tag $y_{i-1}$ to the current tag $y_i$. Adding the CRF layer is helpful to cultivate strong correlations that exist between labels of adjacent neighborhoods.

\section{Evaluation}\label{sec:5}
\subsection{Datasets}
The CleanEval dataset \cite{baroni2008cleaneval} is a cross-domain benchmark, consisting of 55 webpages for training (development) and 676 webpages for evaluation. The ``clean text file" for each webpage does not align properly between text blocks of the two files,  and so cannot be used directly for analysing HTML files at the tag level. We adopt Vogels et al.'s method \cite{vogels2018web2text} to make proper alignment.
%
The latest GoogleTrends-2017 dataset \cite{leonhardt2020boilerplate} published in 2020, consists of 180 documents randomly sampled from a larger pool of websites retrieved from the top Google queries with the corresponding main text. 
It is customary to use 80 webpages for training and the remaining 100 webpages for testing when using GoogleTrends-2017. 

These two datasets, however, are insufficient to train a neural network.
%
It is also more desirable to have a training dataset  
mixing of past and contemporary webpages from a wider range of domains. To this end, we combine the following three published type-1 datasets: 
(1) the dataset 
annotated by Uzun E. et al. \cite{uzun2013hybrid}, consisting of 1,170 webpages from 573 domains 
published from 2008 to 2011; (2) the dataset annotated by Peters M. and Lecocq D. \cite{peters2013content}, consisting of 1,381 webpages collected in late 2012 from RSS feeds, news articles, and blogs; (3) the TECO Benchmark Suite, presented by Alarte J. \cite{alarte2015temex}, consisting of webpages collected from 130 domains during the period from 2013 to 2020. TECO is a multilingual dataset made for testing language independent features and we remove non-English and non-type-1 webpages from it. We 
train SemText on this combined dataset, then fine-tune it using the 55 training webpages in CleanEval and 80 webpages in GoogleTreands-2017. This allows us to demonstrate the robustness of SemText on webpages of different structures across a wider spectrum. 

We also evaluate SemText on 250 type-2 webpages 
consisting of 70 pages sampled from the TECO Forum dataset with non-English pages removed\footnote{This dataset is available at \url{https://github.com/dreamlegends/Semtext}.}, and 
180 webpages collected from 
nine large Q\&A sites including Quora and Yahoo Answer, with main content extracted by custom-built spiders.

\subsection{Model Setup}

Let $n=50$ be the maximum number of words for each word-string component (without stopwords).
%
For each text block $x$, we translate HTML tags in $\text{tag}(x)$ using the W3C HTML5 reference\footnote{https://dev.w3.org/html5/html-author/}. Class names in class($x$) are processed as described in Section \ref{sec:3.2}. While we have not seen any text block $x$ with tag$(x)$ or class$(x)$ containing more than 50 words, it is common for text$(x)$ to exceed the 50-word limit and so truncation is used to trim it down to 50.
With an average of 10 to 15 words per sentence in the main text (with stopwords removed), 50 words can cover 4--5 sentences, sufficient for representing the semantics of a text block. 

Let $m=85$ be the upper bound for the number of text blocks produced by SCS, which is adequate to cover 90\% of news articles and would not cause excessive computation.
If SCS produces more than 85 text blocks for an HTML file, we divide them into sub-sequences in the same order as evenly as possible so that the number of text blocks in each sub-sequence is the largest possible bounded above by 85. 

We use kernels of sizes of 3, 5, and 7 for the word-level CNN encoder, with the corresponding numbers of filters being 128, 128, and 256, for a total number of filters $\ell = 512$. This setting covers consecutive words in almost every aspect to capture the essential meaning of a text block. 
There are 512 hidden units for a single Bi-LSTM. Parameter optimization is performed using stochastic gradient descent (SGD) with a batch size of 64.

We divide at random the combined dataset with a 75-25 split into a training set and a validation set, and choose the checkpoint with the highest F1 score on the validation set. 
Our model is implemented on Pytorch and trained on a single NVIDIA GeForce GTX 2080Ti GPU. 

\subsection{Comparison results}

We compare SemText with BoilerNet \cite{leonhardt2020boilerplate}, Web2Text \cite{vogels2018web2text}, and BoilerPipe \cite{kohlschutter2010boilerplate},
the best models so far,
under the measures of average precision and recall. 
To provide fair comparison, we train and test these models 
as we train SemText using the same combined dataset and fine-tune them
using the same development data of CleanEval and GoodTrends (except BoilerPipe that
is not built for fine-tuning).
We name the models trained this way as, respectively,
BoilerNet-C, Web2Text-C, and BoilerPipe-C. 
We also train these models using the original development data of CleanEval
to obtain, respectively, three models named BoilerNet-1, Web2Text-1, and
BoilerPipe-1; and using the original development data of GoogleTrends to
obtain, respectively, three models named BoilerNet-2, Web2Text-2, and BoilderPipe-2.

We test all models on the same test data of CleanEval and GoogleTrends.
Evaluation is carried out at the text-block level and blocks are treated equally with the same weight. 
Table \ref{tab:1} shows the evaluation results,
where
1$|$2 means that the results of the corresponding model version 1 (e.g., BoilerPipe-1) are under
the CleanEval column and version 2 (e.g., BoilerPipe-2) are
under the GoogleTrends column. The numbers in boldface are the highest.


\begin{center}
\begin{table}[h]
\centering
\caption{Comparison results} 
\begin{tabular}{l|c|c|c||c|c|c}
\hline \multirow{2}{*}{\bf Methods} & \multicolumn{3}{c||}{\textbf{CleanEval}} & \multicolumn{3}{c}{\textbf{GoogleTrends}} \\ 
\cline{2-7}     & \bf ~P~ & \bf ~R~ & \bf ~F1 & \bf ~P~ & \bf ~R~ & \bf ~F1\\ \hline
BoilerPipe-C           & 0.81 	& 0.67 		& 0.73  &0.74 	& 0.61 		& 0.69 \\
Web2Text-C		                & 0.80  & 0.76      & 0.78	&0.76 	& 0.73 		& 0.74 \\
BoilerNet-C		                                &0.88  &0.85  &0.86     &0.87 	& \textbf{0.86}		& 0.87\\
\hline
BoilerPipe-1$|$2		           & 0.87 	& 0.73 		& 0.79  &0.80 	& 0.66 		& 0.72 \\
Web2Text-1$|$2		            & 0.85  & 0.82      & 0.83	&0.82 	& 0.74 		& 0.78 \\
BoilerNet-1$|$2		                                &0.85  &0.80  &0.82     & 0.86  & 0.82      & 0.84\\
\hline
\textbf{SemText}                             &\textbf{0.91} 		& \textbf{0.89}		& \textbf{0.90}   &\textbf{0.92} 	& \textbf{0.86} 		& \textbf{0.89} \\
\hline
\end{tabular}
\label{tab:1}
\end{table}
\end{center}

Indications of these results are summarized below:
\begin{enumerate}
\item 
SemText outperforms 
the previous models under each
category 
except that
SemText and BoilerNet-C have the same recall on GoogleTrends.
%

\item BoilerNet-C is substantially better
than BoilerPipe-C and Web2Text-C on both datasets.
\item BoilerPipe-C and Web2Text-C on GoogleTrends are substantially worse than themselves
on CleanEval. Likewise, BoilerPipe-2 and Web2Text-2 are substantially worse
than BoilerPipe-1 and Web2Text-1, respectively.
\item Web2Text-1 is slightly better than BoilerNet-1, and Web2Text-2 is substantially
lower than BoilerNet-2. 
\item The results of SemText and BoilerNet are consistent on both datasets, but the results of BoilerPipe and Web2Text are not.
\end{enumerate}
A possible cause of inconsistency of BoilerPipe and Web2Text across different datasets is that some of their handcrafted features 
are sensible to webpage structure.
%
For example, 
both web structure and tag usage have changed gradually in the past ten years,
and placing ads from a sidebar to the main content area can increase the link density and decrease the text density, making the text-density feature fail on contemporary webpages. Tag TD, which was widely used 
before 2009, was replaced by tag DIV in the web evolution. Thus, BoilerPipe and Web2Text 
would provide anticipated performance only when 
they are trained and tested on datasets from the same period. 
BoilerNet, on the other hand, does benefit from the combined dataset, which indicates
that neural network models can indeed provide a better and more robust representation. The combined dataset also provides a large vocabulary size to alleviate the Out-of-Vocabulary problem mentioned in the last paragraph of Section \ref{sec:2}. Since each tag is treated as a token in the BoilerNet, the change of tag usage may still affect the model. SemText mitigates this drawback by taking advantage of the semantics registered inside the tags.

\begin{figure}[h]
\centering
\includegraphics[width=0.45\textwidth, height=2.5in]{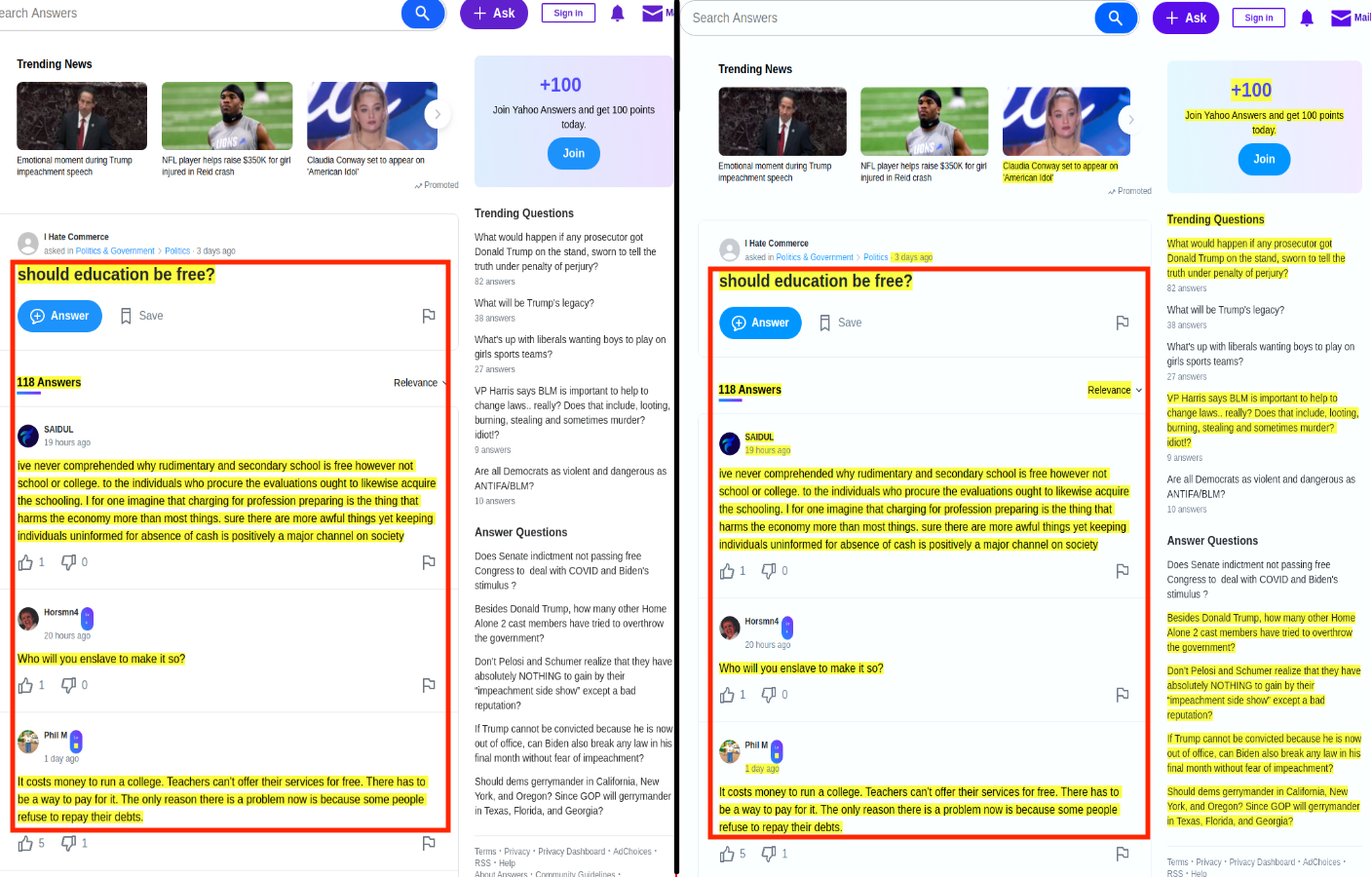} 

\medskip 
\begin{tabular}{cc}
(a) SemText-1 \hspace{0.1in} & \hspace{0.25in} (b) BoilerNet-C1
\end{tabular}
\caption{A type-2 page, where the highlighted texts enclosed in red boxes are the corresponding main content} 
\label{fig:5}
\end{figure}

SemText also performs reasonably well on type-2 webpages without any training on type-2 webpages and is substantially better than BoilerNet-C,
which in turn is substantially better than Web2Text-C and BoilerPipc-C (see the upper part of Table \ref{tab:3}).
\begin{center}
\begin{table}[h]
\centering
\caption{Comparison results on the Type-2 dataset}
\begin{tabular}{l|c|c|c}
\hline \bf Methods & \bf ~P~ & \bf ~R~ & \bf ~F1 \\ \hline
BoilerPipe-C            & \textbf{0.93} &0.12 &0.21	\\
Web2Text-C              &0.59   &0.73 &	0.65 \\
BoilerNet-C             &0.66   & 0.81       &0.72	\\
\textbf{SemText}        &0.79		& \textbf{ 0.84}		& \textbf{0.82}   \\
\hline
BoilerPipe-C1          &\textbf{0.92} &0.19 &0.32 \\
Web2Text-C1	         &0.63 &0.75 &0.68	\\
BoilerNet-C1           &0.72 &\textbf{0.87} &0.79	\\
\textbf{SemText-1 }    & 0.85	&\textbf{0.87}		& \textbf{0.86}   \\
\hline
\end{tabular}
\label{tab:3}
\end{table}
\end{center}
Next, we retrain all models by including type-2 webpages in the training dataset and
evaluate how well they perform on type-2 pages.
To this end, we randomly select 50 webpages from the type-2 dataset and
add them to the combined dataset of type-1 pages. 
The remaining 200 type-2 webpages are used for evaluation.
We obtain SemText-1, BoilerNet-C1, BoilerPipe-C1, and Web2Text-C1, with
the evaluation results shown in the lower part of
Table \ref{tab:3}.
It can be seen that all models have improved performance with the
same performance ranking as before.
In particular, under the F1 measure, SemText-1 is 8.87\% higher than BoilerNet-C1.
While SemText-1 and BoilerNet-C1 have the same recall, SemText-1's precision is much higher.
As shown in Fig \ref{fig:5}, BoilerNet-C1 mistakenly labels the question list in the sidebar and part of the text in the header as the main content, and SemText-1 successfully removes the side bar and the header completely. We observe that the mislabeled text is similar to the main content. The improvement of SemText can be attributed to adding class info (e.g. 
question list, sidebar) to the representation of the question-list text blocks in the side bar.
BoilerPipe has extremely high precision with extremely low recall because BoilerPipe only extracts the first block of the main content.



\subsection{Ablation study}
Let SemText\_TXT denote the baseline of SemText with only text sequence, Semtext\_TAG = SemText\_TXT plus tag sequence, and SemText\_CLS =  SemText\_TXT plus class sequence. Evaluation results are shown in Table \ref{tab:4}.
\begin{center}
\begin{table}[h]
\centering
\caption{Ablation results under GoogleTrends-2017}
\begin{tabular}{l|c|c|c}
\hline 
{\bf Methods}   & \bf ~P~ & \bf ~R~ & \bf ~F1\\ \hline
SemText\_TXT    &0.82	& \textbf{0.87}		& 0.84\\
SemText\_TAG	&0.90 	& 0.84 		& 0.87 \\
SemText\_CLS    &0.89 	& 0.85		& 0.87\\
\textbf{SemText}       &\textbf{0.92} 	& 0.86 		& \textbf{0.89} \\
\hline
\end{tabular}
\label{tab:4}
\end{table}
\end{center}
We can see that SemText\_TXT has the lowest F1 score, while SemText\_TAG and
SemText\_CLS each improves the F1 score over SemTexT\_TXT, indicating that adding a tag sequence and a class sequence for a text block each contributes to the improvement of the baseline model.
A possible cause of the lowest precision score of SemText\_TXT is that it tends to mislabel text blocks that are similar to the main content. On the other hand, SEMText\_TXT has a higher recall, which is
expected. Adding a tag sequence and a class sequence both
contribute to a higher precision, as indented; but they also slightly decrease the recall scores.

%

In summary,  SemText consistently achieves substantially higher F1 scores across type-1 and type-2 webpages over all the evaluated models. Moreover, all models, after trained, run about the same time on the evaluation data. In particular, it takes an average of 38 ms for SemText to extract the main text on a single NVIDIA GeForce GTX 2080Ti GPU, with an average of 683 DOM nodes per page in the evaluation data.

\section{Conclusion and Final Remarks}\label{sec:6}
We have shown that using the semantic representation of text blocks proposed in the paper and classifying them by a hierarchical neural network model is promising on boilerplate detection. 
It would be interesting to explore if this approach may be beneficial to other applications dealing with webpages.

Analyzing the labeling results, we find that mislabeled blocks by SemText often appear at the beginning of a sub-sequence. This phenomenon has a natural explanation: A text block without labels of previous text blocks would  more likely be misclassified, and occurs more often on the type-2 dataset. Thus, it would be interesting to investigate how to 
label the ``entire sequence'' of text blocks without needing to break it into sub-sequences. 

A long sequence is unavoidable on 
community-based Q\&A webpages. On the other hand, the sequencing-labeling model 
confines the length of an input sequence for effective training, and we break a long sequence into sub-sequences somewhat arbitrarily to meet this requirement. Considering that we have already achieved the highest F1 scores over previous methods even with such a simple strategy, it would make sense to investigate how to make use of labels generated from a previous sub-sequence. For example, we may directly use a few text blocks in the previous sequence with their generated labels. An obstacle in this approach is that some of these labels may be incorrect. Thus, we would need to figure out how many previous text blocks should be used to minimize the negative impact of incorrect labels. This would likely become a combinatorial pursuit similar to those for achieving fault tolerance. We may also break a long sequence of text blocks into overlapping sub-sequences. In this direction we would need to investigate 
how much overlap would be appropriate and how to resolve conflicting labels  on overlapped text blocks.

Finally, we may explore a tree-structured LSTM-neural-network model to label text blocks. This approach might be more appropriate for labeling sibling blocks that have the same structure with the same look. 
This direction would likely require more sophisticated modeling efforts.

The code for SemText is available at \url{https://github.com/dreamlegends/Semtext}.
\bibliographystyle{IEEEtran}
\bibliography{sample-base}

\end{document}